\documentclass[letterpaper,journal]{IEEEtran}
\usepackage{amsmath,amsfonts,amssymb}
\usepackage[noend]{algorithmic}
\usepackage{algorithm}
\usepackage{float}
\usepackage{array}
\usepackage[caption=false,font=normalsize,labelfont=sf,textfont=sf]{subfig}
\usepackage{textcomp}
\usepackage{stfloats}
\usepackage{url}
\usepackage{verbatim}
\usepackage{graphicx}
\usepackage{cite}
\usepackage{booktabs} 
\usepackage{multirow}
\usepackage{xcolor}
\usepackage[super]{nth}
\usepackage{tikz}

\usepackage{soul}
\usepackage{enumitem}

\usepackage[switch]{lineno}
\providecommand{\DIFdelbegin}{\protect\color{red}}
\providecommand{\DIFdelend}{\protect\color{black}}

\begin{document}

%%%%%%%%%%%%%%%%%%%%%%%%%%%%%%%%%%%%%%%%%%%%%%%%%%%%%%%%%%%%%%%%%%%%%%%%%%%%%%%%%%%%%%%
%%%%%%%%%%%%%%%%%%%%%%%%%%%%%%%%%%%%%%%%%%%%%%%%%%%%%%%%%%%%%%%%%%%%%%%%%%%%%%%%%%%%%%%

\title{AQ4SViT: An Automated Quantization Framework with Search Gating Policy for Compressing Spiking Vision Transformers}

\author{Rachmad Vidya Wicaksana Putra$^*$,~\IEEEmembership{Member,~IEEE,} Saad Iftikhar$^*$, and Muhammad Shafique,~\IEEEmembership{Senior Member,~IEEE} 
\thanks{Rachmad Vidya Wicaksana Putra and Saad Iftikhar are with eBRAIN Lab, Division of Engineering, New York University (NYU) Abu Dhabi, United Arab Emirates (UAE); (e-mail: rachmad.putra@nyu.edu, si2356@nyu.edu). \\
\indent Muhammad Shafique is the Director of eBRAIN Lab, Division of Engineering, New York University (NYU) Abu Dhabi, United Arab Emirates (UAE); 
(e-mail: muhammad.shafique@nyu.edu). \\ 
\indent $^*$ Equal contributions.}% <-this % stops a space
%\thanks{Manuscript received Month DD, 2025; revised Month DD, 2025.}
}
 \maketitle

%%%%%%%%%%%%%%%%%%%%%%%%%%%%%%%%%%%%%%%%%%%%%%%%%%%%%%%%%%%%%%%%%%%%%%%%%%%%%%%%%%%%%%%
%%%%%%%%%%%%%%%%%%%%%%%%%%%%%%%%%%%%%%%%%%%%%%%%%%%%%%%%%%%%%%%%%%%%%%%%%%%%%%%%%%%%%%%

\begin{abstract}
Spiking Vision Transformers (SViTs) have emerged as alternative low-power ViT models, but their large sizes hinder their deployments on resource-constrained embedded AI systems.  
To address this, state-of-the-art works proposed quantization techniques to compress SViT models, but their manual, human-guided approach needs a huge design time and power/energy consumption to find the appropriate quantization setting for each given network, making this approach not scalable for quantizing multiple networks.
Toward this, we propose \textbf{AQ4SViT}, a novel automated quantization framework for SViTs that can provide quick quantization settings with good trade-offs between accuracy and memory. 
To achieve this, AQ4SViT employs the following key ideas: 
(1) \textit{quantization search strategy} that evaluates the quantization setting candidates while considering the accuracy constraint; and (2) \textit{search gating policy} that quickly evaluates and selects promising quantization candidates by leveraging membrane potential drift as a performance proxy. 
In the search gating policy, AQSViT employs two search algorithm variants to provide trade-off options: \textit{Greedy search}, which performs fast but may lead to local optima; and \textit{Beam search}, which performs slower but has better performance in finding global optima selection due to a wider search space. 
Experimental results show that AQ4SViT-Greedy quickly finds the appropriate quantization settings, achieving up to 6.6$\times$ faster search time and up to 82.5\% memory saving compared to the state-of-the-art; 
while AQ4SViT-Beam further reduces the memory footprint by up to 90\% compared to the state-of-the-art, but with 4.5$\times$ longer search time; 
all these results are obtained while maintaining high accuracy within 1.5\% from the original/non-quantized models on the ImageNet dataset.
These results highlight that our AQ4SViT framework offers substantial advancements toward efficient SViT deployments on resource-constrained embedded AI systems.
\end{abstract}

\begin{IEEEkeywords}
Neuromorphic Computing, Spiking Neural Networks (SNNs), Spiking Vision Transformers (SViTs), Automated Quantization, SViT Model Compression, Search Gating Policy, Embedded AI Systems.
\end{IEEEkeywords}

%%%%%%%%%%%%%%%%%%%%%%%%%%%%%%%%%%%%%%%%%%%%%%%%%%%%%%%%%%%%%%%%%%%%%%%%%%
%%%%%%%%%%%%%%%%%%%%%%%%%%%%%%%%%%%%%%%%%%%%%%%%%%%%%%%%%%%%%%%%%%%%%%%%%%
\section{Introduction}
\label{Sec_Intro}

%%%
\begin{figure}[t]
    \centering
    \includegraphics[width=\linewidth]{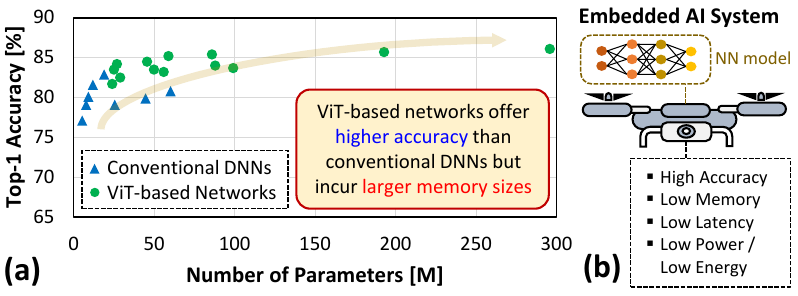}
    \vspace{-0.6cm}
    \caption{\textbf{(a)} ViT models surpass the accuracy of conventional DNNs for a classification task on the ImageNet dataset~\cite{Ref_Deng_ImageNet_CVPR09}, but incur larger memory sizes; based on studies in~\cite{Ref_Han_SurveyViT_TPAMI22}. 
    \textbf{(b)} Embedded AI typically demands for high accuracy as well as low memory, latency, and power/energy consumption~\cite{Ref_Minhas_SurveyNCL_Access25}.}
    \label{Fig_DNNvsVIT}
    \vspace{-0.5cm}
\end{figure}
%%%

Transformer-based networks~\cite{Ref_Vaswani_Attention_NIPS17} have demonstrated state-of-the-art performance in addressing diverse machine learning (ML) tasks.
Recently, vision-based tasks (such as image classification, object detection, and targeted segmentation) have become prominent applications that significantly benefit from Transformers; also known as \textit{Vision Transformers (ViTs)}~\cite{Ref_Han_SurveyViT_TPAMI22}\cite{Ref_Dosovitskiy_Transformers_ICLR21}\cite{Ref_Khan_SurveyViT_CSUR22}.
This trend is showcased in Fig.~\ref{Fig_DNNvsVIT}(a) as ViT models have surpassed the accuracy of state-of-the-art deep neural networks (DNNs).
Therefore, deploying ViTs in resource-constrained embedded AI systems is in high demand and actively being pursued, as this may enable diverse application use-cases that can increase human productivity~\cite{Ref_Han_SurveyViT_TPAMI22}\cite{Ref_Dosovitskiy_Transformers_ICLR21}\cite{Ref_Khan_SurveyViT_CSUR22}.
Despite their potentials, implementing ViTs in resource-constrained systems remains a non-trivial challenge due to their huge network sizes and computational complexity, which lead to significantly large memory cost and high power/energy consumption. 

In the last decade, advancements in spiking neural networks (SNNs) have provided energy-efficient computation model to execute neural network (NN) algorithms, due to their sparse spike-driven operations~\cite{roy2019towards, Ref_Rathi_SNNsurvey_CSUR23, Ref_Putra_FSpiNN_TCAD20, Ref_Bartolozzi_EmbodiedNeuroIntel_Nature22, Ref_Putra_SNNonCNP_IJCNN25, Ref_Putra_SpikeNAS_TAI26, Ref_Putra_QSLM_DATE26}.
Therefore, researchers recently proposed Spiking Vision Transformers (SViTs) as low-power alternatives to ViT models, such as Spikformer with 66.3M parameters~\cite{Ref_Zhou_Spikformer_ICLR23}, Spike-Driven Transformer (SDT) with 66.3M parameters~\cite{Ref_Yao_SpikeDrivenTransformer_NeurIPS23}, Spike-Driven Transformer v2 (SDTv2) with 55.4M parameters~\cite{Ref_Yao_SpikeDrivenTransformer2_ICLR24}, and Spike-Driven Transformer v3 (SDTv3) with 173M parameters~\cite{Ref_Yao_SDTv3_TPAMI25}. Here, M denotes million (10$^6$) of counts. 
However, their large sizes hinder their deployability for resource-constrained embedded AI. 

To reduce the size of SNN-based models, existing works have proposed different techniques in the literature, such as weight pruning~\cite{Ref_Rathi_PruneQuantizeSNN_TCAD18}, quantization~\cite{Ref_Rathi_PruneQuantizeSNN_TCAD18, Ref_Sorbaro_OptimSNN_FNINS20, Ref_Zou_MedianQuant_ISCAS20, Ref_Putra_QSpiNN_IJCNN21}, and neuron elimination~\cite{Ref_Putra_FSpiNN_TCAD20}.
Here, quantization is one of the prominent techniques as it effectively reduces model size at the cost of acceptable accuracy degradation due to information loss.
Existing DNN quantization frameworks cannot be used directly in SNN domain due to substantial differences in the neuronal and synaptic operations.
Besides, SNN quantization requires a careful design to avoid significant accuracy drop~\cite{Ref_Putra_QSpiNN_IJCNN21}.  
Moreover, manually devising an appropriate quantization setting for any given SViT model is not a scalable approach, since it requires huge design time and incurs large power/energy consumption accordingly for each investigated network model.
Therefore, \textit{\textbf{the targeted research problem} in this work is how can we quickly quantize any given SViT models to reduce their memory requirements, while maintaining high accuracy?}
A solution to this problem may enable an efficient design automation for efficient SViT deployments on embedded AI systems; see Fig.~\ref{Fig_DNNvsVIT}(b).

%%%%%%%%%%%%%%%%%%%%%%%%%%%%%%
\subsection{State-of-the-Art and Their Limitations}
\label{Sec_Intro_SOTA}

Currently, most of existing works still target for developing SViT models that achieve high accuracy~\cite{Ref_Zhou_Spikformer_ICLR23, Ref_Yao_SpikeDrivenTransformer_NeurIPS23, Ref_Yao_SpikeDrivenTransformer2_ICLR24}. 
Therefore, only a few state-of-the-art works target SViT model compression. 
For instance, work of~\cite{Ref_Xu_TrimSViT_arxiv24} proposes search with layer-wise quantization in training strategy, thus requiring huge design time and power/energy consumption during its search-and-train process.
Moreover, it still considers relatively small datasets (e.g., N-Caltech101, DVS-Gesture, and CIFAR10-DVS).
Another work, the quantized spike-driven transformer (QSDT)~\cite{Ref_Qiu_QSDT_ICLR25} leverages PTQ and then distillation using information from the artificial neural network (ANN) counterpart for addressing the accuracy degradation from its PTQ stage. 
Hence, this approach is only suitable for SViT models that have established ANN counterparts, and the additional distillation step itself incurs significant training-time overhead (multiple retraining epochs with the ANN teacher), making it inapplicable to task-specific or novel SNN designs where no pre-trained ANN equivalent exists.
Meanwhile, the quantized spiking vision transformer (QSViT)~\cite{Ref_Putra_QSViT_IJCNN25} employs layer-wise PTQ based on the analysis of layer-wise quantization. 
In general, despite their benefits, these state-of-the-art methods are not scalable for quantizing different SViT models due to their \textit{manual} design approaches, i.e., they require human-guided iterative analysis between quantization steps (examining results, hypothesizing new settings, and re-running experiments), leading to long design time and huge power/energy consumption during the optimization process.
\textit{Therefore, an alternative quantization approach is required to quickly quantize any given SViTs in a fully automated, end-to-end manner, while preserving high accuracy}.

%%%%%%%%%%%%%%%%%%%%%%%%%%%%%%
\subsection{Case Study and Related Research Challenges}
\label{Sec_Intro_Challenges}

\begin{figure}[t]
    \centering
    \includegraphics[width=\linewidth]{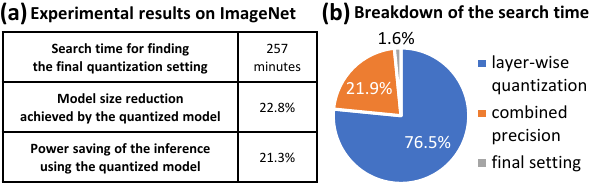}
    \vspace{-0.6cm}
    \caption{\textbf{(a)} Experimental results from quantizing SDTv2 model using the QSViT methodology on the Nvidia GeForce RTX 4090 GPU. \textbf{(b)} Breakdown of the search time.}
    \label{Fig_Observe}
    \vspace{-0.3cm}
\end{figure}

To demonstrate the limitations of state-of-the-art, we perform a case study. 
Specifically, we follow PTQ approach from QSViT~\cite{Ref_Putra_QSViT_IJCNN25}, and evaluate its design/search time to find the quantization setting for compressing the pre-trained SDTv2 model~\cite{Ref_Yao_SpikeDrivenTransformer2_ICLR24} considering the ImageNet dataset through following key steps.
\begin{itemize}[leftmargin=*]
    \item First, we apply layer-wise quantization 
    to generate different quantization settings\footnote{Each quantization setting refers to a set of all precision levels across different layers of the network.},   
    then run the inference for each setting considering all 50,000 (50K) validation images.  
    \item Second, we identify new quantization settings, 
    that potentially lead to acceptable accuracy based on analysis in the first step.
    Then, we run the inference for each setting.
    \item Third, we define the final quantization setting that maintains high accuracy with the most memory saving based on analysis in the second step.
    Then, we run the inference considering all 50K validation images. 
\end{itemize}
To support this, we employ the experimental setup that will be further discussed in Section~\ref{Sec_EvalMethod}.
The experimental results are provided in Fig.~\ref{Fig_Observe}, from which we make the following key observations.
\begin{itemize}[leftmargin=*]
    \item Searching time of the manual quantization approach is relatively long, and mostly dominated by the process for investigating the impact of layer-wise quantization.
    \item Complexity of network architecture (e.g., number of layers) contributes to the number of search iterations during experimental analysis (e.g., through layer-wise quantization and combined precision settings).
    \item Evaluation strategy (e.g., utilizing all 50K validation images) contributes to the evaluation time for each quantization setting, which accumulates over search iterations.    
\end{itemize}
These points expose the following research challenges.
\begin{itemize}[leftmargin=*]
    \item Quantization process should be automated to reduce the burden of design/search time, hence enabling scalable approach for quantizing different possible models.
    \item Quantization process should efficiently handle different possible network complexity (e.g., number of layers).
    \item Quantization process should quickly yet effectively evaluate each quantization setting candidate. 
\end{itemize}

%%%
\begin{figure*}[t]
    \centering
    \includegraphics[width=0.95\linewidth]{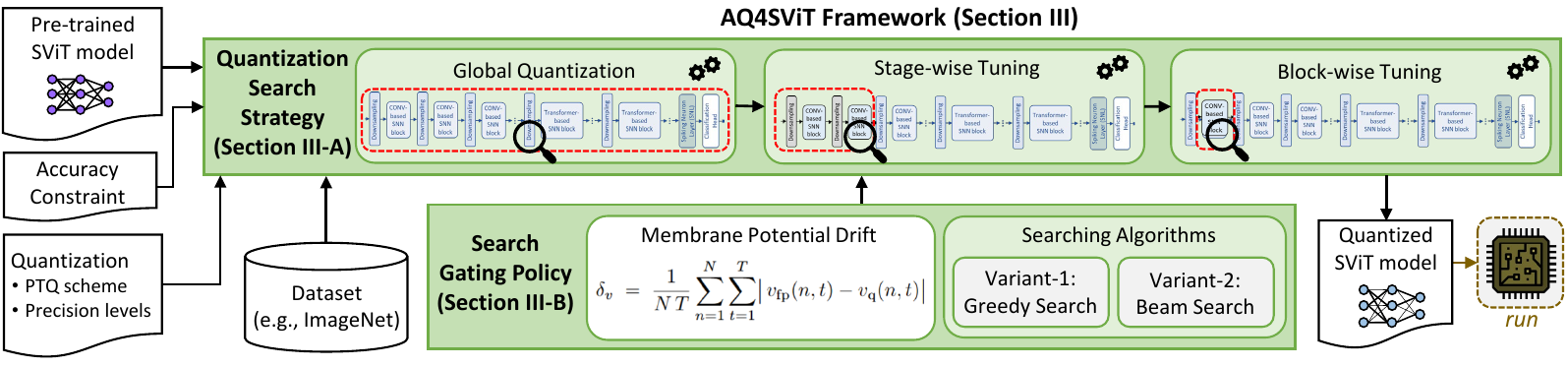}
    \vspace{-0.3cm}
    \caption{The AQ4SViT framework with its novel optimization steps highlighted in green.}
    \label{Fig_AQ4SViT}
    \vspace{-0.3cm}
\end{figure*}
%%%

%%%%%%%%%%%%%%%%%%%%%%
\subsection{Our Novel Contributions}
\label{Sec_Intro_NovelContrib}

To solve the targeted problem and associated research challenges, we propose \textit{\textbf{AQ4SViT}, a novel automated quantization framework that can quickly provide quantization settings for compressing SViTs with good trade-offs between accuracy and memory}.
This effort is also the first PTQ search framework for SViTs considering large-scale datasets (e.g., ImageNet).
It employs the following key ideas; see the overview in Fig.~\ref{Fig_AQ4SViT}.
\begin{itemize}[leftmargin=*]
    \item \textbf{Quantization Search Strategy:} 
    It aims to evaluate quantization setting candidates under different quantization phases (i.e., global, stage-wise, and block-wise), while considering the accuracy constraint.
    \item \textbf{Search Gating Policy:}
    It aims to evaluate each quantization setting candidate by leveraging \textit{membrane potential drift} as a proxy for accuracy, so that only promising candidates proceed to costly full-dataset evaluation.
    To achieve this, we propose two search algorithm variants that provide trade-offs between search time and compression rate.
    \begin{itemize}
        \item \textit{Greedy search}: It typically performs fast due to aggressive selection, but may settle at local-optima.
        \item \textit{Beam search}: It typically performs slower than the greedy search due to a wider search space, but it can lead to better performance in finding models that have smaller memory footprints.
    \end{itemize}
 \end{itemize}

\textbf{Key Results:} 
We evaluate the AQ4SViT framework using a PyTorch-based implementation and run it on an Nvidia GeForce RTX 4090 24GB GPU machine. 
Experimental results show that, our AQ4SViT can provide the appropriate quantization settings to maintain high accuracy within 1.5\% from the original/non-quantized models on the ImageNet dataset. 
Specifically, AQ4SViT-Greedy achieves this with up to 6.6$\times$ faster search time and up to 82.5\% memory saving compared to the state-of-the-art QSViT~\cite{Ref_Putra_QSViT_IJCNN25}.
Meanwhile, AQ4SViT-Beam achieves this at the cost of 4.5$\times$ longer search time, but enables up to 90\% memory saving compared to the state-of-the-art~\cite{Ref_Putra_QSViT_IJCNN25}.
These results highlight that AQ4SViT framework represents a significant advancement toward efficient SViT deployments on resource-constrained embedded AI systems.

%%%%%%%%%%%%%%%%%%%%%%%%%%%%%%%%%%%%%%%%%%%%%%%%%%%%%%%%%%%%%%%%%%%%%%%%%%%
%%%%%%%%%%%%%%%%%%%%%%%%%%%%%%%%%%%%%%%%%%%%%%%%%%%%%%%%%%%%%%%%%%%%%%%%%%%
\section{Background}
\label{Sec_Back}

%%%%%%%%%%%%%%%%%%%%%%%%%%%%%%
\subsection{Spiking Vision Transformer (SViT)}
\label{Sec_Back_SViT}

SNNs have demonstrated their immense potential as energy-efficient alternatives to NN algorithms~\cite{Ref_Putra_SNNonCNP_IJCNN25}\cite{Ref_Putra_SpikeNAS_TAI26}\cite{Ref_Akopyan_TrueNorth_TCAD15}. 
Hence, they are actively studied for vision transformers, known as Spiking Vision Transformers (SViTs).
Currently, in the literature, state-of-the-art SViT models include Spikformer \cite{Ref_Zhou_Spikformer_ICLR23}, Spikingformer~\cite{Ref_Zhou_Spikingformer_AAAI26}, Spike-Driven Transformer (SDT)~\cite{Ref_Yao_SpikeDrivenTransformer_NeurIPS23}, Spike-Driven Transformer v2 (SDTv2)~\cite{Ref_Yao_SpikeDrivenTransformer2_ICLR24}, and Spike-Driven Transformer v3 (SDTv3)~\cite{Ref_Yao_SDTv3_TPAMI25}.
These models have similar main architectural components, such as \textit{spiking convolutional (CONV) layers}, \textit{spiking linear layers}, and \textit{spiking attention layers}.
Since these models have deep structure and multiple layers may contribute to a specific functionality, their layers are often grouped into a \textit{block}, and multiple blocks are grouped into a \textit{stage}. 
Furthermore, these models usually also employ a downsampling block. 
For instance, SDTv2 model has 4 stages and a classification head.
In stage-1, there are 2 sets of downsampling and CONV blocks. 
In stage-2, there is a downsampling block, which is followed by 2 CONV blocks. 
In stage-3, there is a downsampling block, which is followed by 6 attention blocks. 
In stage-4, there is a downsampling block, which is followed by 2 attention blocks.

%%%%%%%%%%%%%%%%%%%%%%%%%%%%%%
\subsection{Quantization Methods}
\label{Sec_Back_QuantMethod}

There are two possible schemes for quantizing SNN models, namely \textit{Quantization-aware Training (QAT)} and \textit{Post-Training Quantization (PTQ)}~\cite{Ref_Putra_QSpiNN_IJCNN21}\cite{Ref_Krishnamoorthi_Whitepaper_arXiv18}.
QAT applies quantization on an SNN model during the training phase, hence resulting in a quantized SNN model that is ready for inference phase. 
Meanwhile, PTQ applies quantization on a pre-trained SNN model, hence resulting in a quantized SNN model for inference.  
In this work, we use PTQ since it does not need costly training, hence providing a low-cost quantization approach, which is suitable for enabling efficient search algorithms. 
In this work, we convert the floating-point 32bit (FP32) values into the targeted bit precision ($b$) by identifying the conversion scale ($\mu$) based on the range of weight values ($w_{range}=w_{max}-w_{min}$) from the floating-point format and the range of quantized values ($Q_{range}=Q_{max}-Q_{min}$); see Eq.~\ref{Eq_Scale}.
Here, $Q_{max} = 2^{(b-1)}-1 $ and $Q_{min} = -2^{(b-1)}$.
Once the scale factor $\mu$ is defined, each weight ($w$) can be quantized into its integer format ($w_q$). We realize this conversion through simulated quantization to enable fast exploration, while providing representative accuracy and memory information~\cite{Ref_vanBaalen_SimQuantRealPower_CVPR22}.
\begin{equation}
\begin{split}
   \mu = \frac{w_{range}}{Q_{range}} \;\;\;\; \text{and} \;\;\;\; w_q = \text{round} \left( \frac{w}{\mu} \right) 
\end{split}
\label{Eq_Scale}
\end{equation} 

%%%%%%%%%%%%%%%%%%%%%%%%%%%%%%%%%%%%%%%%%%%%%%%%%%%%%%%%%%%%%%%%%%%%%%%%%%%
%%%%%%%%%%%%%%%%%%%%%%%%%%%%%%%%%%%%%%%%%%%%%%%%%%%%%%%%%%%%%%%%%%%%%%%%%%%
\section{The AQ4SViT Framework}
\label{Sec_Method}

This framework considers the following key steps: \textit{quantization search strategy} and \textit{search gating policy} as described below (see an overview in Fig.~\ref{Fig_AQ4SViT}).

%%%%%%%%%%%%%%%%%%%%%%%%%%%%%%
\vspace{-0.2cm}
\subsection{Quantization Search Strategy}
\label{Sec_Method_Search}

We define the automated searching pipeline that facilitates different precision levels across network layers, as effective quantization settings may have heterogeneous precision levels due to variability of layers' sensitivity~\cite{Ref_Putra_QSViT_IJCNN25}.  
Here, we consider two search algorithm variants to provide trade-off options.
\begin{itemize}[leftmargin=*]
    \item \textit{Greedy Search (AQ4SViT-Greedy)}: 
    It aims to provide a fast solution by leveraging its aggressive search and selection, but it may settle for local-optima.
    \item \textit{Beam Search (AQ4SViT-Beam)}: 
    It aims to provide better search results than the greedy search by exploring a wider heuristic search space for a stronger memory-saving solution, but it may perform slower than the greedy search. 
\end{itemize}

\smallskip
\subsubsection{\textbf{AQ4SViT-Greedy}} 
The following are its key steps (see the pseudo-codes in Alg.~\ref{Alg_SearchStrategy}). 

%%%
\begin{algorithm}[t]
\caption{AQ4SViT-Greedy Search Pipeline}
\label{Alg_SearchStrategy}
\begin{algorithmic}[1]
    \footnotesize
    \renewcommand{\algorithmicrequire}{\textbf{INPUT:}}
    \renewcommand{\algorithmicensure}{\textbf{OUTPUT:}}
    \REQUIRE
        Pre-trained FP32 model ($\mathcal{M}$); Quantized model ($M_q$);
        Pre-defined bit precision levels $G=\{16,12,8,4\}$;
        Training set ($\mathcal{C}$); Validation set ($\mathcal{V}$); Timesteps ($T$);
        Sample size for gating ($N_{\mathrm{met}}=128$); Accuracy tolerance ($\Delta_{\mathrm{acc}}$, in top-1 \%-age);
        Drift threshold ($\tau_v$); 
        Accuracy ($A$); Memory ($M$); Minimum bits ($b_{\min}=3$);
    \ENSURE
        Selected bit precision setting ($\mathcal{B}^*$); \\
    \smallskip
    \textbf{Initialization:}
        \STATE $A_{\mathrm{fp}},M_{\mathrm{fp}}$ = evaluate($\mathcal{M}$, $\mathcal{V}$);
        \STATE $\mathcal{B}[:,:]$ = 32; $\mathcal{R}$ = []; \\
    \smallskip
    \textbf{Process:} \\
    \underline{// Step-1: Global Quantization}
      \FOR{$b\in G$}
        \STATE $\mathcal{M}_q$ = quantize$(\mathcal{M},b)$;
        \STATE $\delta_v$ = gating$(\mathcal{M},\mathcal{M}_q,\mathcal{C},N_{\mathrm{met}},T)$; // Alg.~\ref{Alg_SearchGating}
        \IF{$\delta_v\le\tau_v$}
          \STATE $A,M$ = evaluate$(\mathcal{M}_q,\mathcal{V})$;
        \IF{$A \ge A_{\mathrm{fp}}-\Delta_{\mathrm{acc}}$}
          \STATE $\mathcal{R}$ = append($\mathcal{R}$, record$(:, :, b, A, M, \mathrm{valid})$); $\mathcal{B}[:,:]$ = $b$;
        \ENDIF
      \ENDIF
    \ENDFOR
    \smallskip
    \underline{// Step-2: Stage-wise Tuning}
    \FOR{each stage $s$}
      \STATE $\mathrm{hi}$ = $\mathcal{B}[s,:]$; $\mathrm{lo}$ = $4$;
      \WHILE{($\mathrm{hi}$ - $\mathrm{lo}>1$)}
        \STATE $c$ = $\lfloor(\mathrm{hi}+\mathrm{lo})/2\rfloor$; $\mathcal{B}[s,:]$ = $c$;
        \STATE $\mathcal{M}_q$ = quantize$(\mathcal{M},\mathcal{B})$;
        \STATE $\delta_v$ = gating$(\mathcal{M},\mathcal{M}_q,\mathcal{C},N_{\mathrm{met}},T)$; // Alg.~\ref{Alg_SearchGating}
        \IF{$\delta_v\le\tau_v$}
          \STATE $A,M$ = evaluate$(\mathcal{M}_q,\mathcal{V})$;
          \IF{$A \ge A_{\mathrm{fp}}-\Delta_{\mathrm{acc}}$}
            \STATE $\mathcal{R}$ = append($\mathcal{R}$, record$(s,:,c, A, M, \mathrm{valid})$; $\mathrm{hi}$ = $c$;
          \ELSE
            \STATE revert $\mathcal{B}$; $\mathrm{lo}$ = $c$;
          \ENDIF
        \ELSE
          \STATE revert $\mathcal{B}$; $\mathrm{lo}$ = $c$;
        \ENDIF
      \ENDWHILE
    \ENDFOR
    \smallskip
    \underline{// Step-3: Block-wise Tuning}
    \FOR{each block $k$ with $\mathcal{B}[:,k]>4$}
      \STATE $b_{\mathrm{k}}$ = $\max(\lceil\mathcal{B}[:,k]/2\rceil, b_{\min})$ ; $\mathcal{B}[:,k]$ = $b_{\mathrm{k}}$;
      \STATE $\mathcal{M}_q$ = quantize$(\mathcal{M},\mathcal{B})$;
      \STATE $\delta_v$ = gating$(\mathcal{M},\mathcal{M}_q,\mathcal{C},N_{\mathrm{met}},T)$; // Alg.~\ref{Alg_SearchGating}
      \IF{$\delta_v\le\tau_v$}
        \STATE $A,M$ = evaluate$(\mathcal{M}_q,\mathcal{V})$;
        \IF{$A \ge A_{\mathrm{fp}}-\Delta_{\mathrm{acc}}$}
          \STATE $\mathcal{R}$ = append($\mathcal{R}$, record$(:,k,b_k, A, M, \mathrm{valid})$;
        \ELSE
          \STATE revert $\mathcal{B}$;
        \ENDIF
      \ENDIF
    \ENDFOR
    \STATE $\mathcal{B}^* \leftarrow$ valid trial in $\mathcal{R}$ with minimum memory $M$; 
    \smallskip
    \RETURN $\mathcal{B}^*$;
\end{algorithmic}
\linenumbers
\end{algorithm}
\DIFdelbegin
\DIFdelend \setlength{\textfloatsep}{6pt}
%%%

\begin{itemize}[leftmargin=*]
    \item \textbf{Step-1. Global Quantization:}
    In this step, we quantize all weight parameters across all layers with the same precision level to quickly identify the minimum uniform precision level that meets the given accuracy constraint.
    It includes the following points.
    \begin{itemize}
        \item The precision levels are pre-defined and stored in list $G$, which are explored one-by-one for quantization; see Alg.~\ref{Alg_SearchStrategy}: lines 3-9.
        \item Once the model is quantized, we perform \textit{search gating}\footnote{Details of the \textit{search gating function} and its criteria will be discussed in Section~\ref{Sec_Method_Gating}.} to quickly evaluate if the investigated quantization setting forms a network that meets the criteria; see Alg.~\ref{Alg_SearchStrategy}: lines 5-6.
        \item If the accuracy is acceptable, the investigated quantization setting is stored in $\mathcal{B}$; see Alg.~\ref{Alg_SearchStrategy}: lines 7-9.
        \item Once the loop is finished, we move to Step-2.
    \end{itemize}
    \item \textbf{Step-2. Stage-wise Tuning:}
    In this step, we partition the network into stages and perform search for finding the appropriate precision level for each stage; see Alg.~\ref{Alg_SearchStrategy}: lines 10-23.
    It includes the following points.
    \begin{itemize}
        \item We employ binary search based on the precision from step-1 (global quantization) and the lowest pre-defined precision (e.g., 4 bit); see Alg.~\ref{Alg_SearchStrategy}: lines 11-13.
        Crucially, the binary search explores \emph{intermediate} bit-width values (e.g., 6-bit or 7-bit) between the global precision and the minimum, enabling finer-grained mixed-precision assignments beyond the discrete set $G$.
        \item Once the model is quantized, we perform \textit{search gating} to quickly evaluate if the investigated quantization setting forms a network that meets the criteria; see Alg.~\ref{Alg_SearchStrategy}: lines 15-16.
        \item If the observed accuracy is acceptable, then the investigated quantization setting in $\mathcal{B}$ is kept, otherwise $\mathcal{B}$ is reverted back; see Alg.~\ref{Alg_SearchStrategy}: lines 17-23.
        \item Once the loop is finished, we move to Step-3.
    \end{itemize}
    \item \textbf{Step-3. Block-wise Tuning:}
    In this step, we partition each stage into blocks, and perform search for finding the appropriate precision level for each block; see Alg.~\ref{Alg_SearchStrategy}: lines 24-33.
    It includes the following points.
    \begin{itemize}
        \item We gradually reduce the precision level by half at a time if the current bit-width is larger than 4 bits; see Alg.~\ref{Alg_SearchStrategy}: lines 24-25. 
        The global and stage-wise searches use 4 bits as the minimum precision, while the block-wise step may go to 3 bits for selected blocks only.
        \item Once the model is quantized, we perform \textit{search gating} to quickly evaluate if the investigated quantization setting meets with the criteria; see Alg.~\ref{Alg_SearchStrategy}: lines 28-29.
        \item If accuracy is acceptable, then the investigated quantization setting in $\mathcal{B}$ is kept, otherwise $\mathcal{B}$ is reverted back; see Alg.~\ref{Alg_SearchStrategy}: lines 30-33. 
        \item Once the loop is finished, we will obtain the appropriate quantization setting that meets the accuracy constraint, which is recorded in $\mathcal{B}$; see Alg.~\ref{Alg_SearchStrategy}: line 34. 
    \end{itemize}
\end{itemize}

\smallskip
\subsubsection{\textbf{AQ4SViT-Beam}} 
Its key idea is to expand every parent configuration into a set of children at each step, evaluate all children, 
and retain only the top-$K$ valid candidates for the next step.
A dedicated \textit{Repair-pass} further refines each beam member by attempting local precision adjustments at the block level.
The terminal selection criterion is minimum parameter memory subject to the accuracy constraint. 
Its pseudo-code is shown in Alg.~\ref{Alg_BeamSearch} and its key steps are described below.

\setlength{\textfloatsep}{4pt plus 1pt minus 1pt}
\begin{algorithm}[!t]
\nolinenumbers
\caption{AQ4SViT-Beam Search Pipeline}
\label{Alg_BeamSearch}
\begin{algorithmic}[1]
  \footnotesize
  \renewcommand{\algorithmicrequire}{\textbf{INPUT:}}
  \renewcommand{\algorithmicensure}{\textbf{OUTPUT:}}
  \REQUIRE
    Pre-trained FP32 model ($\mathcal{M}$);
    Pre-defined bit precisions $G=\{16,12,8,4\}$; Training set ($\mathcal{C}$);
    Validation set ($\mathcal{V}$); Timesteps ($T$);
    Sample size ($N_{\mathrm{met}}=128$); Accuracy tolerance ($\Delta_{\mathrm{acc}}$, in top-1 \%-age);
    Drift threshold $\tau_v$; Beam size ($K=4$);
    Minimum bits ($b_{\min}=3$);
  \ENSURE 
    elected bit precision setting ($\mathcal{B}^*$); \\
  \smallskip
  \textbf{Initialization:} \\
  \STATE $A_{\mathrm{fp}}, M_{\mathrm{fp}}$ = evaluate($\mathcal{M}$, $\mathcal{V}$); Beam = $\emptyset$; \\
  \smallskip
  \textbf{Process:} \\
  \underline{// Step-1: Global Quantization} \\
  \FOR{$b \in G$ \textbf{(descending)}}
    \STATE $\mathcal{M}_q$ = quantize($\mathcal{M}$, $b$);
    \STATE $\delta_v$ = gating($\mathcal{M}$, $\mathcal{M}_q$, $\mathcal{C}$, $N_{\mathrm{met}}$, $T$); // Alg.~\ref{Alg_SearchGating}
    \IF{$\delta_v \le \tau_v$}
      \STATE $A$, $M_{\mathrm{sz}}$ = evaluate($\mathcal{M}_q$, $\mathcal{V}$);
      \IF{$A \ge A_{\mathrm{fp}}-\Delta_{\mathrm{acc}}$}
        \STATE Beam = Beam $\cup$ \{record($b$, $A$, $M_{\mathrm{sz}}$, valid)\};
      \ENDIF
    \ENDIF
  \ENDFOR
  \STATE Beam = \textsc{top}$_K$(Beam, $K$); \\
  \smallskip
  \underline{// Step-2: Stage-wise Tuning} \\ 
  \FOR{each stage $s$}
    \STATE Children = $\emptyset$;
    \FOR{each parent $\in$ Beam}
      \STATE Children = Children $\cup$ \{parent.clone()\}; // keep current
      \FOR{each $b' \in \mathrm{adj\_lower}(\mathrm{parent.bits}[s],\,G,\,b_{\min})$}
        \STATE child = parent.clone(); child.bits$[s]$ = $b'$;
        \STATE $\mathcal{M}_q$ = quantize($\mathcal{M}$, child.bits);
        \STATE $\delta_v$ = gating($\mathcal{M}$, $\mathcal{M}_q$, $\mathcal{C}$, $N_{\mathrm{met}}$, $T$); // Alg.~\ref{Alg_SearchGating}
        \IF{$\delta_v \le \tau_v$}
          \STATE $A$, $M_{\mathrm{sz}}$ = evaluate($\mathcal{M}_q$, $\mathcal{V}$);
          \IF{$A \ge A_{\mathrm{fp}}-\Delta_{\mathrm{acc}}$}
            \STATE Children = Children $\cup$ \{child\};
          \ENDIF
        \ENDIF
      \ENDFOR
    \ENDFOR
    \STATE Beam = \textsc{top}$_K$(Children, $K$);
  \ENDFOR
  \smallskip
  \underline{// Step-3: Block-wise Tuning}
  \FOR{each block $k$}
    \STATE Children = $\emptyset$;
    \FOR{each parent $\in$ Beam}
      \STATE Children = Children $\cup$ \{parent.clone()\};
      \STATE $b'$ = $\max(\lceil\mathrm{parent.bits}[k]/2\rceil,\,b_{\min})$;
      \STATE child = parent.clone(); child.bits$[k]$ = $b'$;
      \STATE $\mathcal{M}_q$ = quantize($\mathcal{M}$, child.bits);
      \STATE $\delta_v$ = gating($\mathcal{M}$, $\mathcal{M}_q$, $\mathcal{C}$, $N_{\mathrm{met}}$, $T$); // Alg.~\ref{Alg_SearchGating}
      \IF{$\delta_v \le \tau_v$}
        \STATE $A$, $M_{\mathrm{sz}}$ = evaluate($\mathcal{M}_q$, $\mathcal{V}$);
        \IF{$A \ge A_{\mathrm{fp}}-\Delta_{\mathrm{acc}}$}
          \STATE Children = Children $\cup$ \{child\};
        \ENDIF
      \ENDIF
    \ENDFOR
    \STATE Beam = \textsc{top}$_K$(Children, $K$);
  \ENDFOR
  \smallskip
  \underline{// Step-4: Repair Pass}
  \FOR{each $c \in$ Beam}
    \FOR{each block $k$ (descending by $c$.bits$[k]$)}
      \STATE child = $c$.clone(); child.bits$[k]$ = adj\_lower($c$.bits$[k]$, $G$, $b_{\min}$)[0]; 
      \STATE $\mathcal{M}_q$ = quantize($\mathcal{M}$, child.bits); 
      \STATE $\delta_v$ = gating($\mathcal{M}$, $\mathcal{M}_q$, $\mathcal{C}$, $N_{\mathrm{met}}$, $T$); // Alg.~\ref{Alg_SearchGating}
      \IF{$\delta_v \le \tau_v$ \textbf{and} valid(child) \textbf{and} child.$M_{\mathrm{sz}} < c.M_{\mathrm{sz}}$}
        \STATE $c \leftarrow$ child;
      \ENDIF
    \ENDFOR
    \FOR{each block $k$ (ascending by $c$.bits$[k]$)}
      \STATE child = $c$.clone(); child.bits$[k]$ = adj\_higher($c$.bits$[k]$, $G$);
      \STATE $\mathcal{M}_q$ = quantize($\mathcal{M}$, child.bits);
      \STATE $\delta_v$ = gating($\mathcal{M}$, $\mathcal{M}_q$, $\mathcal{C}$, $N_{\mathrm{met}}$, $T$); // Alg.~\ref{Alg_SearchGating}
      \IF{$\delta_v \le \tau_v$ \textbf{and} valid(child) \textbf{and} improves\_acc\_valid(child, $c$)}
        \STATE $c \leftarrow$ child;
      \ENDIF
    \ENDFOR
  \ENDFOR
  \smallskip
  \underline{// Step-5: Terminal Selection}
  \STATE $\mathcal{B}^* \leftarrow \arg\min_{c \in \mathrm{Beam},\, c.\mathrm{valid}}\; c.M_{\mathrm{sz}}$;
  \RETURN $\mathcal{B}^*$;
\end{algorithmic}
\end{algorithm}

\begin{itemize}[leftmargin=*]
    \item \textbf{Step-1. Global Quantization:}
    Each uniform precision level $b \in G$ is applied across all layers and evaluated through the gating function; see Alg.~\ref{Alg_BeamSearch}: lines 2-9 for global quantization and Alg.~\ref{Alg_SearchGating} for the gating function.
    Candidates that pass the gate and meet the accuracy constraint are added to the initial beam, which is then pruned to the top-$K$ by proxy score.
    \item \textbf{Step-2. Stage-wise Tuning:}
    For each stage $s$, every beam member (parent) spawns children by retaining the current precision or attempting adjacent lower precision levels for that stage; see Alg.~\ref{Alg_BeamSearch}: lines 10-22.
    Each new child is evaluated through the gating function. 
    Children that pass the gate and meet the accuracy constraint are added to the candidate pool.
    Then, the beam is updated with the top-$K$ candidates.
    \item \textbf{Step-3. Block-wise Tuning:}
    We reduce the precision of individual blocks by halving the value; see Alg.~\ref{Alg_BeamSearch}: lines 23-35.
    Similar to AQ4SViT-Greedy, the global and stage-wise searches operate down to 4 bits, while the block-wise step may reduce selected blocks one level further down to 3 bits.
    Again, only children that pass the gate and meet accuracy constraints can enter the candidates in the beam.
    \item \textbf{Step-4. Repair Pass:}
    For each candidate $c$ in the beam, we attempt to push individual blocks to a lower precision or higher precision and check if it recovers accuracy while keeping the candidate valid); see Alg.~\ref{Alg_BeamSearch}: lines 36-48.
    Each attempted change is first evaluated through the drift gate before full accuracy evaluation.
    \item \textbf{Step-5. Terminal Selection:}
    The final configuration $\mathcal{B}^*$ is the valid candidate in the beam with the minimum parameter memory; see Alg.~\ref{Alg_BeamSearch} \end{itemize}

\begin{algorithm}[t]
\caption{Search Gating Function}
\label{Alg_SearchGating}
\begin{algorithmic}[1]
  \footnotesize
  \renewcommand{\algorithmicrequire}{\textbf{INPUT:}}
  \renewcommand{\algorithmicensure}{\textbf{OUTPUT:}}
  \REQUIRE Pre-trained FP32 model ($\mathcal{M}$); Quantized model ($\mathcal{M}_q$);
     Training set ($\mathcal{C}$); Sample size ($N_{\mathrm{met}}$); Timesteps ($T$);
  \ENSURE 
    Membrane potential drift ($\delta_v$); \\
  \smallskip
  \textbf{Process:} \\ 
  \underline{// Data Collection}
  \STATE $v_{\mathrm{fp}},\; \omega$ $\leftarrow$ run$(\mathcal{M},\, \mathcal{C},\, N_{\mathrm{met}})$;
         // membrane potentials, spike weights
  \STATE $v_{\mathrm{q}}$ $\leftarrow$ run$(\mathcal{M}_q, \mathcal{C},\, N_{\mathrm{met}})$; \\
  \smallskip
  \underline{// Membrane Potential Drift}
  \STATE $S_{\mathrm{dri}} \leftarrow 0$; \quad $S_{\omega} \leftarrow 0$;
  \FOR{each spiking layer $l$}
    \STATE $T' \leftarrow \min(T_{\mathrm{fp}}^{(l)}, T_{\mathrm{q}}^{(l)})$; // align temporal lengths
    \STATE $\delta_v^{(l)} \leftarrow \dfrac{1}{N^{(l)}\,T'}
           \displaystyle\sum_{n,t}\bigl|v_{\mathrm{fp}}^{(l)}(n,t) - v_{\mathrm{q}}^{(l)}(n,t)\bigr|$;
    \STATE $S_{\mathrm{dri}} \mathrel{+}= \omega^{(l)} \cdot\, \delta_v^{(l)}$;
           \quad $S_{\omega} \mathrel{+}= \omega^{(l)}$;
  \ENDFOR
  \STATE $\delta_v \leftarrow S_{\mathrm{dri}} / \max(S_{\omega}, \varepsilon)$; // mass-weighted network-level drift
  \RETURN $\delta_v$;
\end{algorithmic}
\end{algorithm}

%%%%%%%%%%%%%%%%%%%%%%%%%%%%%%
\subsection{Search Gating Policy}
\label{Sec_Method_Gating}

We have identified that evaluating quantization candidates is the main design-time bottleneck from the state-of-the-art; see Fig.~\ref{Fig_Observe}.
Toward this, \textit{we propose a novel \textbf{search gating policy} to quickly yet effectively evaluate if the given quantization setting is likely to achieve high accuracy}.
It applies a lightweight filtering criterion and a small batch of samples ($N_{\mathrm{met}}$) to the quantized model for predicting if the model is likely to achieve high accuracy.

\smallskip
\textbf{Membrane Potential Drift ($\delta_{v}$):}
It aims to identify if neurons in the quantized model behave similarly to those in the original non-quantized model.
It is useful because even when a quantized model produces the same spike counts, its neurons' membrane potentials may drift from the original non-quantized model, leading to different spike timings and hence causing accuracy degradation.
\textit{We define the layer-wise membrane potential drift ($\delta_v^{(l)}$) as the mean absolute error between the membrane potential sequences of the FP32 and quantized models at layer $l$, and aggregate across layers using a spike-count mass-weighting scheme}; see Eq.~\ref{Eq_Drift}.
Here, $v_{\mathrm{fp}}^{(l)}(n,t)$ and $v_{\mathrm{q}}^{(l)}(n,t)$ are the membrane potential of neuron $n$ at timestep $t$ for the FP32 and quantized models, respectively, $T' = \min(T_{\mathrm{fp}}^{(l)}, T_{\mathrm{q}}^{(l)})$ aligns the temporal sequences, and $\omega^{(l)}$ is the spike-count activity weight at layer $l$.
\begin{equation}
  \begin{split}
    \delta_v^{(l)} & = \frac{1}{N^{(l)}\,T'}
    \sum_{n=1}^{N^{(l)}}\sum_{t=1}^{T'}
    \bigl|\,v_{\mathrm{fp}}^{(l)}(n,t) - v_{\mathrm{q}}^{(l)}(n,t)\bigr|, \\
    \delta_v & = \frac{\sum_l \omega^{(l)}\,\delta_v^{(l)}}{\sum_l \omega^{(l)}}
  \end{split}
  \label{Eq_Drift}
\end{equation}
A candidate passes the drift gate if $\delta_v \le \tau_v$, where $\tau_v$ denotes the drift threshold.

\begin{figure}[t]
    \centering
    \includegraphics[width=\linewidth]{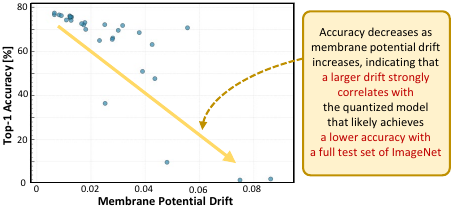}
    \vspace{-0.6cm}
    \caption{Accuracy of different models with different quantization settings and membrane potential drift scores.}
    \label{Fig_Acc_vs_Drift}
    \vspace{0.4cm}
\end{figure}

\smallskip
\textbf{Correlation with Accuracy:}
To validate that $\delta_v$ is an effective surrogate for accuracy, we evaluate multiple quantized SDTv2 models with different quantization settings on the full ImageNet validation set; results are shown in Fig.~\ref{Fig_Acc_vs_Drift}.
Accuracy decreases monotonically as $\delta_v$ increases, with a Spearman rank correlation of $\rho=-0.959$ ($p=2\times10^{-5}$), confirming that smaller drift strongly predicts that the quantized model will likely achieve higher accuracy on a full test evaluation.

\smallskip
\textbf{Sensitivity Analysis of $\tau_v$:}
The drift threshold $\tau_v$ controls the selectivity of the gating policy.
To validate that our chosen value is robust and not over-fit to a specific model, we conduct a systematic sweep of $\tau_v$ on SDTv2; results are summarized in Table~\ref{Tab_TauSweep}.
The results show that $\tau_v=0.0136$ achieves the best trade-off, as it leads to notable reduction in evaluated trials (from 29 to 7) while maintaining Top-1 accuracy (i.e., 77.96\%) within 1\% accuracy from the baseline non-quantized model (i.e., 78.9\%).
Thresholds below 0.0136 become too conservative.
We also observe similar patterns for SDT and SDTv3 models, thereby confirming that the chosen $\tau_v=0.0136$ is positioned in the stable operating region and generalizes across SDT, SDTv2 and SDTv3 architectures without per-model tuning.
\begin{table}[h]
  \centering
  \small
  \setlength{\tabcolsep}{4pt}
  \caption{Sensitivity of drift threshold $\tau_v$ on SDTv2 obtained using our AQ4SViT\_Greedy}
  \label{Tab_TauSweep}
    \begin{tabular}{c c c c}
      \toprule
      $\tau_v$ & Trials & Trial Reduction [\%] & Top-1 [\%] \\
      \midrule
      $\infty$ (no gating) & 29 & 0.0   & 77.96 \\
      0.050               & 25 & 13.8  & 77.96 \\
      0.030               & 19 & 34.5  & 77.96 \\
      0.020               & 13 & 55.2  & 77.96 \\
      0.015               & 8 & 72.4  & 77.96 \\
      \textbf{0.0136}      & \textbf{7}  & \textbf{75.9} & \textbf{77.96} \\
      0.010               & 6 & 79.3  & 78.16 \\
      0.005               & 5 & 82.8  & 78.49 \\
      \bottomrule
    \end{tabular}
    \smallskip
\end{table}

\smallskip
\textbf{Policy:}
Our proposed search gating is implemented as a function presented in Alg.~\ref{Alg_SearchGating}.
A quantization candidate is \emph{gate-passed} if $\delta_v \le \tau_v$, and only gate-passed candidates proceed to full-dataset evaluation.
Therefore, candidates with high membrane potential drift (indication of corrupted spike dynamics) are efficiently eliminated early, dramatically curtailing the number of costly full-dataset evaluations; see Alg.~\ref{Alg_SearchStrategy}: lines 6, 16, and 28 for greedy-based search, and Alg.~\ref{Alg_BeamSearch}: lines 7, 18, 31, 41, and 47 for beam-based search.

%%%%%%%%%%%%%%%%%%%%%%%%%%%%%%%%%%%%%%%%%%%%%%%%%%%%%%%%%%%%%%%%%%%%%%%%%%%
\section{Evaluation Methodology}
\label{Sec_EvalMethod}

To evaluate our AQ4SViT framework, we build the experimental setup based on the PyTorch-based implementation (i.e., PyTorch~1.12 and CUDA~11.6), that employs the SpikingJelly library~\cite{Ref_Fang_SpikingJelly_SciAdv23}. 
Afterward, we run it on a single Nvidia RTX 4090 24GB GPU device with Ubuntu OS v22.04.
This setup takes a pre-trained SViT model, post-training quantization (PTQ) scheme, pre-defined precision levels (i.e., 16, 12, 8, and 4 bits), accuracy constraint (i.e., maximum 1.5\% accuracy degradation), and ImageNet-1K dataset~\cite{Ref_Deng_ImageNet_CVPR09} as inputs. 
We consider the state-of-the-art SDT~\cite{Ref_Yao_SpikeDrivenTransformer_NeurIPS23}, SDTv2~\cite{Ref_Yao_SpikeDrivenTransformer2_ICLR24}, and SDTv3~\cite{Ref_Yao_SDTv3_TPAMI25} as the pre-trained SViT models.
We run SDT, SDTv2, and SDTv3 using their open-source codes in the same GPU machine to provide fair comparison, and we obtain 74.06\% accuracy for SDT, 78.9\% accuracy for SDTv2, and 86.4\% accuracy for SDTv3, which are slightly different from data in the original papers (i.e., 74.57\% for SDT, 79.7\% for SDTv2, and 86.2\% for SDTv3).
The drift threshold is set as $\tau_v=0.0136$ based on our sensitivity analysis (Table~\ref{Tab_TauSweep}) due to its stable operating point.
We use $T=4$ timesteps for SDT and SDTv2, and employ $T=8$ timesteps for SDTv3, following the default settings from the original papers~\cite{Ref_Yao_SpikeDrivenTransformer_NeurIPS23}\cite{Ref_Yao_SpikeDrivenTransformer2_ICLR24}\cite{Ref_Yao_SDTv3_TPAMI25}. 
For the comparison partner, we consider the state-of-the-art QSViT~\cite{Ref_Putra_QSViT_IJCNN25} as it does not require additional operations, such as costly re-training or distillation.
The outputs of experiments include the searching time for obtaining the quantization setting for compressing SViT models, top-1 accuracy when running the quantized SViT models, and memory footprints.
 
\begin{table*}[t]
  \centering
  \caption{Comparison of quantization methods for SViT models on ImageNet-1K. \\
  (*) Accuracy from our experiments using the open-source code from the original authors.}
  \label{Tab_Results}
  \small
  \setlength{\tabcolsep}{2.5pt}
  \begin{tabular}{l c c c c c c c c}
    \toprule
    \textbf{Work}
      & \textbf{Backbone }
      & \textbf{\#Params.}
      & \textbf{Precision}
      & \textbf{Memory}
      & \textbf{Memory Saving}
      & \textbf{Timestep}
      & \textbf{Top-1 Accuracy}
      & \textbf{Search Time} \\
    \textbf{}
      & \textbf{Architecture}
      & \textbf{[M]}
      & \textbf{[bit]}
      & \textbf{[MB]}
      & \textbf{[\%]}
      & \textbf{[$T$]}
      & \textbf{[\%]}
      & \textbf{[min.]} \\
    \midrule
    \multicolumn{9}{l}{\textbf{\textit{Baselines (FP32)}}} \\
    Spikformer~\cite{Ref_Zhou_Spikformer_ICLR23}
      & Spikformer & 66.3  & 32 & 253       & ---  & 4 & 74.81  & --- \\
    SDT~\cite{Ref_Yao_SpikeDrivenTransformer_NeurIPS23}
      & SDT & 66.3  & 32 & 253       & ---  & 4 & 74.06* & --- \\
    SDTv2~\cite{Ref_Yao_SpikeDrivenTransformer2_ICLR24}
      & SDTv2 & 55.4  & 32 & 211       & ---  & 4 & 78.90* & --- \\
    SDTv3~\cite{Ref_Yao_SDTv3_TPAMI25}
      & SDTv3 & 173 & 32 & 692 & --- & 8 & 86.40* & --- \\
    \midrule
    \multicolumn{9}{l}{\emph{\textbf{State-of-the-art Quantization for SViTs}}} \\
    QSDT~\cite{Ref_Qiu_QSDT_ICLR25} 
      & SDTv2 &  6.8 & 4     & 3.2  & 98.5 & 4 & 80.30 & ANN distillation \\
    QSViT~\cite{Ref_Putra_QSViT_IJCNN25}
      & SDTv2 & 55.4 & mixed & 163  & 22.8 & 4 & 76.80 & 257 \\
    \midrule
    \multicolumn{9}{l}{\textit{\textbf{AQ4SViT-Greedy (Ours) --- Greedy search}}} \\
    AQ4SViT-Greedy 
      & SDT & 66.3  & mixed %  
      & 77 & 69.6 & 4 & 73.20 & 98 \\
    AQ4SViT-Greedy 
      & SDTv2 & 55.4  & mixed 
      & 63 & 70.1 & 4 & 77.96 & 39 \\
    AQ4SViT-Greedy 
      & SDTv3 & 173 & mixed 
      & 121 & 82.5 & 8 & 85.45 & 127 \\
    \midrule
    \multicolumn{9}{l}{\textit{\textbf{AQ4SViT-Beam (Ours) --- Beam search}}} \\ 
    AQ4SViT-Beam
      & SDT & 66.3 & mixed 
      & 33 & 87 & 4 & 73.90 & 2853 \\
    AQ4SViT-Beam
      & SDTv2 & 55.4 & mixed 
      & 21 & 90 & 4 & 77.54 & 1153 \\
    AQ4SViT-Beam
      & SDTv3 & 173 & mixed 
      & 85 & 87.7 & 8 & 86.22 & 4001 \\
    \bottomrule
  \end{tabular}%
  % }
\end{table*}

%%%%%%%%%%%%%%%%%%%%%%%%%%%%%%%%%%%%%%%%%%%%%%%%%%%%%%%%%%%%%%%%%%%%%%%%%%%
%%%%%%%%%%%%%%%%%%%%%%%%%%%%%%%%%%%%%%%%%%%%%%%%%%%%%%%%%%%%%%%%%%%%%%%%%%%
\section{Results and Discussion}
\label{Sec_Results}

%%%%%%%%%%%%%%%%%%%%%%%%%%%%%%%%%%%%%%%%%%%%%%%
\subsection{Preservation of High Accuracy}
\label{Sec_Results_Accuracy}

Table~\ref{Tab_Results} shows the experimental results for accuracy from baselines, state-of-the-art QSViT, and our AQ4SViT.
These results show that AQ4SViT-Greedy meets with the given accuracy constraint (i.e., maximum 1.5\% accuracy degradation) for all generated quantized models.
For the SDT model, AQ4SViT-Greedy achieves 73.20\% accuracy and AQ4SViT-Beam achieves 73.90\% accuracy, which is within 1.5\% of the baseline non-quantized accuracy of 74.06\%.
For the SDTv2 model, AQ4SViT-Greedy achieves 77.96\% accuracy and AQ4SViT-Beam achieves 77.54\% accuracy, which is within 1.5\% of the baseline non-quantized accuracy of 78.9\%, and competitive against QSDT (80.30\%) which employs expensive distillation from an ANN counterpart.
Meanwhile, for the SDTv3 model, AQ4SViT-Greedy achieves 85.45\% accuracy and AQ4SViT-Beam achieves 86.22\% accuracy, which is within 1.5\% of the baseline non-quantized accuracy of 86.4\%. 
All these results demonstrate that the accuracy constraint is successfully fulfilled across different SViT models.
The reason is that, AQ4SViT-Greedy and AQ4SViT-Beam include the accuracy constraint and employs the search gating policy, which ensure that only the promising quantization setting candidates continue to evaluation with the full dataset.

%%%%%%%%%%%%%%%%%%%%%%%%%%%%%%%%%%%%%%%%%%%
\normalcolor
\subsection{Reduction of Memory Footprints}
\label{Sec_Results_Memory}

Experimental results for the memory footprints from baselines, state-of-the-art QSViT, and our AQ4SViT are shown in Table~\ref{Tab_Results}.
These results show that, both AQ4SViT-Greedy and AQ4SViT-Beam effectively reduce the given SViT models, achieving higher memory saving than the state-of-the-art.
For the SDT model, AQ4SViT-Greedy achieves 69.6\% memory saving, reducing the model from 253~MB to 77~MB, while AQ4SViT-Beam achieves 87\% memory saving, reducing the model from 253~MB to 33~MB
For the SDTv2 model, AQ4SViT-Greedy achieves 70.1\% memory saving (reduction from 211~MB to 63~MB) and AQ4SViT-Beam achieves 90\% memory saving (reduction from 211~MB to 21~MB), which are significantly higher than the state-of-the-art QSViT with 22.8\% memory saving.
Meanwhile, for the SDTv3 model, AQ4SViT-Greedy achieves 82.5\% memory saving (reduction from 692~MB to 121~MB) and AQ4SViT-Beam achieves 87.7\% memory saving (reduction from 692~MB to 85~MB). 
These significant memory savings come from our systematic hierarchical search in AQ4SViT that gradually quantizes and quickly evaluates candidates at network-level, stage-level, and block-level, hence having higher possibility to find quantized models that achieve high accuracy with small memory footprints.
Here, QSDT achieves more memory saving than others, but requires costly distillation and an ANN counterpart to maintain high accuracy, which is infeasible for SViT models without established ANN equivalents.

%%%%%%%%%%%%%%%%%%%%%%%%%%%%%%%%
\normalcolor
\subsection{Searching Time Speed-ups}
\label{Sec_Results_SearchTime}

Experimental results for searching time in Table~\ref{Tab_Results} show that our AQ4SViT effectively expedites the quantization process compared to the state-of-the-art work.
For the SDTv2 model, AQ4SViT-Greedy expedites the search time by 6.6x, since it only requires 39 minutes of search as compared to  the state-of-the-art QSViT with 257 minutes.
For a larger model (i.e., SDTv3), AQ4SViT-Greedy completes the search in 127~minutes, demonstrating that search time scales well relative to model size.
These search time speed-ups come from the following sources.
\begin{itemize}[leftmargin=*]
    \item Quantization search strategy that decomposes the network into hierarchical search spaces (i.e., network-level, stage-level, and block-level, respectively), avoiding costly layer-wise quantization analysis.
    \item Search gating policy, that quickly evaluates the quantization setting candidates through: (1) the employment of mini-batch samples ($N_{\mathrm{met}}{=}128$), which avoids the use of the full dataset; and (2) the drift-based gate, which further curtails the search space by eliminating candidates with degraded spike dynamics early, hence reducing the number of costly full-dataset evaluations.
\end{itemize}
Meanwhile, AQ4SViT-Beam generally incurs a longer search time than AQ4SViT-Greedy and state-of-the-art QSViT due to a wider search space, but it typically leads to more memory savings and often offers better performance (e.g., accuracy).

%%%%%%%%%%%%%%%%%%%%%%%%%%%%%%%%%%%%%%%%%%%%%%%%%%
\subsection{Comparison between AQ4SViT-Greedy and AQ4SViT-Beam}
\label{Sec_Results_Beam}

Table~\ref{Tab_Results} also presents the results of AQ4SViT-Beam on SDT, SDTv2, and SDTv3.
These results show that AQ4SViT-Beam consistently achieves smaller memory footprints than the corresponding AQ4SViT-Greedy configurations, confirming the benefit of maintaining multiple candidate configurations across all search phases.
The beam-guided exploration in AQ4SViT-Beam prevents the premature commitment to sub-optimal stage-wise precision assignments, and the repair pass further refines each beam member toward the memory-efficient solution.
Therefore, AQ4SViT-Beam has a wider exploration space that can uncover configurations for more memory-efficient and possibly higher accuracy than the greedy-based solution. 
The trade-offs between search speed and compression rate demonstrate that AQ4SViT-Greedy and AQ4SViT-Beam offer different benefits, i.e., AQ4SViT-Greedy aims for rapid compression pipelines and AQ4SViT-Beam aims for maximizing memory reduction, while meeting the accuracy constraint.
Such trade-offs (between search time and model compression rate) are important and needed for different compute resources.
 
%%%%%%%%%%%%%%%%%%%%%%%%%%%%%%%%%%%%%%%%%%%%%%%%%%%%%%%%%%%%%%%%%%%%%%%%%%%
%%%%%%%%%%%%%%%%%%%%%%%%%%%%%%%%%%%%%%%%%%%%%%%%%%%%%%%%%%%%%%%%%%%%%%%%%%%
\normalcolor
\section{Conclusion}
\label{Sec_Conclusion}

We propose a novel AQ4SViT framework to automatically find the quantization setting that meets the accuracy constraint, while minimizing the memory footprint.
AQ4SViT employs a hierarchical search strategy with a membrane potential drift gating policy to quickly select promising quantization settings. 
Its search strategy has two variants based on Greedy search (AQ4SViT-Greedy) and Beam search (AQ4SViT-Beam) to offer trade-off between search time and memory compression rate.
Experimental results show that, our AQ4SViT successfully provides the appropriate quantization settings to maintain high accuracy within 1.5\% from the original non-quantized SViT models on the ImageNet-1K dataset. 
AQ4SViT-Greedy achieves this through faster search time (up to 6.6$\times$ faster) and with reduced memory footprint (up to 82.5\% reduction) as compared to the state-of-the-art.
Meanwhile, AQ4SViT-Beam achieves this at the cost of longer search time (4.5$\times$ longer), but enables significant memory reduction (up to 90\% reduction) as compared to the state-of-the-art.
Therefore, our AQ4SViT framework offers an efficient and scalable SViT compression approach, which is a substantial advancement toward enabling efficient SViT deployments for embedded AI.

\section*{Acknowledgment}
This work was partially supported by the NYUAD Center for Artificial Intelligence and Robotics (CAIR), funded by Tamkeen under the NYUAD Research Institute Award CG010.

%%%%%%%%%%%%%%%%%%%%%%%%%%%%%%%%%%%%%%%%%%%%%%%%%%%%%%%%%%%%%%%%%%%%%%%%%%%%%%%%%%
\bibliographystyle{IEEEtran}
\bibliography{bibliography}

\end{document}